\def\year{2022}\relax
\documentclass[letterpaper]{article} 
\usepackage{aaai22}  
\usepackage{times}  
\usepackage{helvet}  
\usepackage{courier}  
\usepackage[hyphens]{url}  
\usepackage{graphicx} 
\usepackage{natbib}  
\usepackage{caption} 
\DeclareCaptionStyle{ruled}{labelfont=normalfont,labelsep=colon,strut=off} 
\usepackage{wrapfig}
\usepackage{subfigure}

\usepackage{algorithm,float}
\usepackage{booktabs} 
\usepackage{bbm}
\usepackage{algorithmicx}
\usepackage[utf8]{inputenc}
\usepackage{makecell}
\usepackage{amsmath}  
\usepackage{lipsum}
\usepackage{subfig}
\usepackage{pbox}
\usepackage{colortbl}
\usepackage{array}
\usepackage{mathtools}
\usepackage{float}
\usepackage{amssymb}
\usepackage{flexisym}
\usepackage{graphicx}
\usepackage{longtable}
\usepackage{multirow}
\usepackage{multicol}
\usepackage{textcomp}
\usepackage{bm}
\usepackage{algpseudocode}
\usepackage{mathrsfs}
\usepackage{comment}

\newcommand{\model}{MTCA}

\definecolor{light-gray}{gray}{0.95}

\usepackage{color}
\usepackage{xcolor}
\definecolor{darkgreen}{rgb}{0,0.3,0}

\DeclareMathAlphabet\mathbfcal{OMS}{cmsy}{b}{n}
\newsavebox\foobox

\usepackage[english]{babel}
\usepackage{blindtext}

\title{Multi-Domain Transformer-Based Counterfactual Augmentation for Earnings Call Analysis}

\author{Zixuan Yuan$^{1}$, Yada Zhu$^{2}$, Wei Zhang$^{3}$, Ziming Huang$^{4}$, Guangnan Ye$^{2}$, Hui Xiong$^{1}$\\}
\affiliations{$^{1}$Rutgers University, $^{2}$IBM Research, $^{3}$Wayfair, $^{4}$Sogou Inc.\\
\{zy101, hxiong\}@rutgers.edu, \{yzhu, gye\}@us.ibm.com, wzhang5@wayfair.com, hzmyouxiang@gmail.com}

\begin{document}
\maketitle
\begin{abstract}
Earnings call (EC), as a periodic teleconference of a publicly-traded company, has been extensively studied as an essential market indicator because of its high analytical value in corporate fundamentals.
The recent emergence of deep learning techniques has shown great promise in creating automated pipelines to benefit the EC-supported financial applications.
However, these methods presume all included contents to be informative without refining valuable semantics from long-text transcript and suffer from EC scarcity issue. Meanwhile, these black-box methods possess inherent difficulties in providing human-understandable explanations. 
To this end, in this paper, we propose a \textit{\underline{M}ulti-Domain \underline{T}ransformer-Based \underline{C}ounterfactual \underline{A}ugmentation}, named \model, to address the above problems.
Specifically, we first propose a transformer-based EC encoder to attentively quantify the task-inspired significance of critical EC content for market inference.
Then, a multi-domain counterfactual learning framework is developed to evaluate the gradient-based variations after we perturb limited EC informative texts with plentiful cross-domain documents, enabling \model~to perform unsupervised data augmentation. 
As a bonus, we discover a way to use non-training data as instance-based explanations for which we show the result with case studies.
Extensive experiments on the real-world financial datasets demonstrate the effectiveness of interpretable \model~for improving the volatility evaluation ability of the state-of-the-art by 14.2\% in accuracy. 
\end{abstract}

\section{Introduction}

Recent years have witnessed a surge of efforts devoted to mining textual data for predicting market movements in the Artificial Intelligence community. 
The vast majority of existing studies~\cite{khedr2017predicting, bing2014public, yang2020html} focus on short and copious texts with dedicated aspects, such as news and twitters.
However, many highly valuable financial documents are long (e.g., $\ge$ 500 words), sparse (e.g., published quarterly or yearly), and contain a broad range of market-concerned topics, including supply chain, operational efficiency, or natural disasters, etc.
It is meaningful yet challenging to cope with these characteristics of financial documents using modern data-intensive neural approaches.

As a typical example of long sparse documents, earnings call (EC) transcript can provide indispensable value in company fundamentals and deliver far-reaching impacts on the financial market. 
Conventionally, market practitioners leverage industrial knowledge to estimate the changes of market uncertainty around earnings forecasts~\cite{rogers2009earnings, de2012bank}.
With the emergence of deep learning, a variety of automated methods~\cite{qin2019you, yang2020html} have been developed to explore salient market signals from EC transcripts.
However, these methods suffer from data scarcity issue, and lack the ability of refining informative contents from lengthy documents to interpret predictions.

To fill in the gap, we follow the concept of counterfactual learning~(CL)~\cite{van2019interpretable}, where the informative EC contents can be identified as potential decision-influencing factors by asking the counterfactual: how would the outcome change if the selected texts were modified? 
Such CL enables us to leverage abundant cross-domain texts (e.g., news release) to perturb EC's informative contents for data augmentation and post-hoc explanation. 
However, applying CL to the existing neural functions is a non-trivial task due to the following challenges.

\noindent \textbf{Complex informativeness modeling.} 
Since the EC-triggered price movement of a company's stock is tightly dependent on informative EC contents, we define the informativeness of these contents as the degree of market influence, which can be characterized by two vital features: domain-topic relevance and task-based significance.
Domain-topic relevance refers to the likelihood of EC contents being labeled by a domain topic.
Different companies' ECs usually reveal unique topics of relevance. 
For instance, the environmental-related topics are crucial discussed points of EC for PG\&E\footnote{https://www.fool.com/earnings/call-transcripts/2021/07/30/pge-corporation-pcg-q2-2021-earnings-call-transcri/} while not so much for companies in the banking industry.
However, it remains unclear how to match textual information with the domain topics.
In addition, task-based significance stands for the importance of an EC towards a particular financial task (e.g., volatility prediction).
The time-evolving nature of financial market makes it complicated to conduct precise significance evaluation and distill redundant features of low-ranked texts. 
Therefore, how to refine salient information from noisy contents with topic awareness is our first challenge.

\noindent \textbf{Unsupervised multi-domain CL.}
Unlike traditional CL that operates on training instances within a single domain, our work targets at using cross-domain unlabeled data from a different source (e.g., financial news) to augment training data (i.e., EC transcripts), which is guided by a post-hoc explanation method under the counterfactual concept, to evaluate the potentials of unlabeled data in optimizing prediction model.
To achieve this goal, we would (i) enforce a high degree of consistency between original and perturbed EC in terms of topic relevance and the task-based significance of the perturbed EC, and (ii) produce the unsupervised labels of perturbed instances to enable sufficient model training using these augmented data.
Therefore, an unsupervised learning paradigm of incorporating the solutions to both requirements poses our second challenge.

To tackle the above challenges, we propose a \textit{\underline{M}ulti-Domain \underline{T}ransformer-Based \underline{C}ounterfactual \underline{A}ugmentation}, named \model, to predict and reason upon market status in a semi-supervised manner. 
Our main contributions are summarized as follows:
\begin{itemize}
\item We develop the transformer-based EC encoder, which applies the hierarchical sparse self-attention to capture precise long-range dependencies among EC sentences, and quantifies the task-oriented importance with top-weighted semantics.
\item We propose the multi-domain counterfactual learning framework to evaluate the gradient-based variations based on cross-domain EC perturbations, which enforces \model~to facilitate unsupervised data augmentation and model explanations.
For model training, we iterate between the supervised update of EC encoder and the unsupervised augmentation of multi-domain texts.
Experimental results on large-scale financial datasets demonstrate the effectiveness and interpretability of \model, by outperforming the state-of-the-art by a large margin of 14.2\% in accuracy. 
\end{itemize}

\section{Related Work}

\indent \textbf{EC-Based Financial Forecasting.}
With the prevalence of natural language processing techniques, EC transcripts have profoundly been exploited to evaluate potential risks in financial markets~\cite{qin2019you, yang2020html, sawhney2020voltage}.
For instance, \cite{qin2019you} proposed a multimodal deep regression model (MDRM) that jointly encodes text and audio information inside EC for financial risk evaluation. 
Recently, \cite{yang2020html} reshaped the MDRM structure with a multi-task hierarchical transformer to predict both short- and long-term variability of market price.
However, we argue that these black-box approaches are difficult to generate trustworthy explanations, making it impractical to be deployed in many safety-critical financial applications.

\noindent \textbf{Sparse Attention.}
Substantial research on efficient sparse attention has been conducted on how to mitigate the quadratic scaling issue of attention being applied to the long-sequence modeling~\cite{tay2020sparse, zhou2021informer, martins2016softmax}.
The general idea behind sparse weights is to enforce the model to only focus on a limited quantity of items at a time.
For instance, \cite{tay2020sparse} developed dynamic sequence truncation method to tailor the Sinkhorn Attention for obtaining efficient quasi-global local attentions.
This can denoise redundant items appeared in the long input sequence and become a useful inductive bias to help discover most influential features as model explanations.
However, it is unsafe to treat all top-weighted items as decisive explanations without any domain guidance, since attention units in the sophisticated networks might demonstrate less fidelity to feature importance measures~\cite{jain2019attention}.

\noindent \textbf{Counterfactual Learning Based Data Augmentation.}
As a sample-based explanation method, counterfactual learning (CL) is designed to evaluate how the model's decision could be altered through minimal changes to the input features~\cite{artelt2019computation}.
This concept has been applied in data augmentation~\cite{pitis2020counterfactual, zmigrod2019counterfactual, wang2020robustness}, which aims to augment real data by performing counterfactual perturbations to a subset of the decisive factors and leaving the rest unchanged.
Recent studies develop the influence function~\cite{koh2017understanding} or its hessian-free variant~\cite{zhang2021sample} to understand the counterfactual effect of training points on a prediction.
To the best of our knowledge, this is the first study that adopts the CL for EC augmentation and model interpretation.

\section{Problem Formulation}

Suppose there are M earning calls $\mathcal{E} = \{E_1, \dots, E_M\}$, each of which is a sequence of sentences padded to the maximum length of $L$. 
We are also accessible to a set of $N$ sentences from another source for EC augmentation, i.e. $\mathcal{S} = \{s_1, \dots, s_N\}$. Throughout the paper, we use boldface lowercase and capital letters (e.g., $\mathbf{e}$ and $\mathbf{E}$) to denote vectors and matrices, and all $\mathbf{W}_s$ and $\mathbf{b}_s$ are represented as the trainable weight matrices and bias terms, respectively.

This paper incorporates cross-domain data to augment EC transcripts for predicting and explaining the future development of financial market.
Specifically, we study the problem of predicting future market volatility, which is regarded as a key indicator of market risk.
For a given stock, we calculate the average  log volatility~\cite{hull2014} over $n$ days following the date of EC. 
We categorize the volatility values into three different levels: downward, steady, and upward fluctuations, labeled as $\{0, 1, 2\}$, respectively.
Here, we formulate this problem as a classification task, with the purpose of learning a mapping function $\mathcal{F}(E) \rightarrow \{0, 1, 2\}$.

\section{Proposed Approach}
\begin{figure}[t]
    \includegraphics[width = 0.45\textwidth]{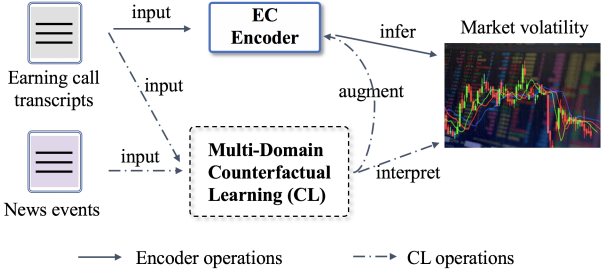}
    \caption{The overview of \model.}
    \label{framework}
\end{figure}

\noindent\textbf{Framework Overview.}
Figure \ref{framework} presents the architecture of our proposed \textbf{\model}, which entails three major parts:
(i) encoding the critical EC texts for market volatility inference,
(ii) generating the multi-domain augmented data and counterfactual explanations,
and (iii) performing the augmentation-enhanced model training in the semi-supervised manner.

In the first part, we develop the transformer-based \textit{EC encoder}, in which the EC texts are encoded into a bag of sentence-level vectors using a pre-trained language model. 
Then, the \textit{Hierarchical sparse self-attention} block is incorporated to precisely capture long-range dependencies among EC sentences for identifying the highly-weighted semantics.
The ultimate EC representations are later fed into a fully-connected layer to output the predicted volatility.

In the second part, we propose the multi-domain counterfactual learning (CL) framework, where we use the expert-tailored topics to train the \textit{Topic classifier} for identifying topic-concerned semantics and establishing semantic connections among multi-domain documents. 
The \textit{Unsupervised counterfactual augmentation} block then incorporates the $\mathrm{TracIn}^+$ function to quantify the gradient-based variations when we randomly perturb meager informative EC contents with abundant cross-domain texts.
This allows \model~to generate sufficient labeled data for the encoder's update, and investigate the task-inspired salience of perturbed semantics for providing the non-training-instance-based (NTIB) interpretation.

In the third part, we introduce a KL-regularized optimization to alternate between the supervised training of the EC encoder and the unsupervised augmentation of multi-domain semantics.

\subsection{EC Encoder}
The transformer-based EC encoder for volatility prediction includes a hierarchical sparse self-attention block and a prediction block.

\subsubsection{Hierarchical Sparse Self-Attention}
Inspired by the Informer \cite{zhou2021informer} that demonstrates the significance of distilling trivial feature maps with sparse attention, we propose the hierarchical sparse self-attention to capture the long-range sentence-level dependencies for EC encoding.
Specifically, it is comprised of two stacked layers of multi-head sparse self-attention and 1-D convolutional filter with the activation function $\mathrm{PeLU}(\cdot)$: 
\small
\begin{eqnarray}
\begin{cases}
\mathbf{\tilde{E}} = SP^{2}(SP^{1}(\mathbf{E}\mathbf{W}_{inp} + \mathbf{b}_{inp})), \\
SP^{i}(\cdot) = \mathrm{MaxPool}(\mathrm{PeLU}(\mathrm{Conv1d}([\cdot]_{HS}))),\\
[\cdot]_{HS} = \mathrm{Concat}(head_1^{i}, \dots, head_h^{i})\mathbf{W}_h^{i}, \\
head_j^{i} = \mathrm{Softmax}(\frac{\mathbf{\hat{Q}}_j^{i}{\mathbf{K}_j^{i}}^\top}{\sqrt{d}})\mathbf{V}_j^{i},
\end{cases}
\label{predictor}
\end{eqnarray}
\normalsize
in which $\mathbf{\tilde{E}}$ is the ultimate representation of EC, and
$\mathbf{W}_{inp} \in \mathbb{R}^{d \times d}$.
For the $i$-th stacked layer $SP^{i}(\cdot)$, $i \in \{1, 2\}$, $\mathbf{\hat{Q}} \in \mathbb{R}^{\frac{d}{i} \times \frac{d}{i}}$,  $\mathbf{K} \in \mathbb{R}^{\frac{d}{i} \times \frac{d}{i}}$, $\mathbf{V} \in \mathbb{R}^{\frac{d}{i} \times \frac{d}{i}}$.
To refine salient semantic features from long-text EC transcript, we design two following functions in Eq.(\ref{predictor}).
First, we add a max-pooling layer $\mathrm{MaxPool}(\cdot)$ with stride $2$ to down-sample $\mathbf{E}$ into its half slice for after stacking one layer.
Second, we develop a sparse matrix $\mathbf{\hat{Q}}$, where we only need to consider
the top-$n_E$ queries under the sparsity estimation $S(\mathbf{q}_x^i, \mathbf{K}_j^{i}) = \max\limits_{y}\{\frac{\mathbf{q}_x {\mathbf{k}_y}^\top}{\sqrt{d/i}}\} - \frac{1}{d/i}\sum_{y=1}^{d/i}\{\frac{\mathbf{q}_x \mathbf{k}_y^\top}{\sqrt{d/i}}\}$, where $\mathbf{q}_x^i$, $\mathbf{k}_y$ stand for the i-th row in the j-th head queries $\mathbf{Q}^{i}_j$ and keys $\mathbf{K}^{i}_j$, respectively. 
By filtering out trivial weights, such max-pooling operation and sparsity matrix enable the encoder to focus on those attentively important texts instead of the redundant ones.
To initialize an EC representation $\mathbf{E} \in \mathbb{R}^{L \times d}$, we encode its $i$-th sentence into the embedding vector $\mathbf{e} = LM(e) + \mathbf{p} \in \mathbb{R}^{1 \times d}$, where $LM(\cdot)$ denotes a sentence-level encoding function of the pre-trained language model, (e.g., RoBERTa~\cite{liu2019roberta})
and we use the identical calculation of the position embedding $\mathbf{p}$ as \cite{vaswani2017attention}:
\begin{eqnarray}
\begin{cases}
\mathbf{p}_{pos, 2m} = sin(pos/10000^{2f/d}),\\
\mathbf{p}_{pos, 2m+1} = cos(pos/10000^{2f/d}),
\end{cases}
\end{eqnarray}
where $pos$ is the sentence position and $f$ is the dimension of sentence embedding.
\subsubsection{Prediction} 
Based on the learned EC representation $\mathbf{\tilde{E}}$, we next adopt a fully-connected layer to forecast the significance level $\mathbf{y}$ of market volatility over $n$-days following the date of EC:
\begin{eqnarray}
\mathbf{y} = \mathrm{Softmax}(\mathbf{W}_p \mathbf{\tilde{E}} + \mathbf{b}_p).
\end{eqnarray}
Note that other sophisticated sequential methods (e.g., LSTM~\cite{hochreiter1997long}) can also be applied here as the prediction function.

\noindent \textit{Remarks.} The above EC encoder has empirically demonstrated strong abilities in modeling hidden task-based significance of informative texts and outperforms the state-of-the-art by a large margin.
However, three crucial issues have yet to be solved: (i) The EC encoder, similar to those in~\cite{yang2020html, sawhney2021empirical}, suffers from the lack of training instances, which may harm their overall generalization.
(ii) The black-box essence of these approaches possesses inherent difficulties in interpreting the predicted volatility.
(iii) It is insecure to treat all top-weighted texts as informative contents or decisive explanations without any domain guidance, since attention units in the sophisticated networks might demonstrate less fidelity to feature importance measures~\cite{jain2019attention}.

\begin{figure}[t]
    \includegraphics[width = 0.48\textwidth]{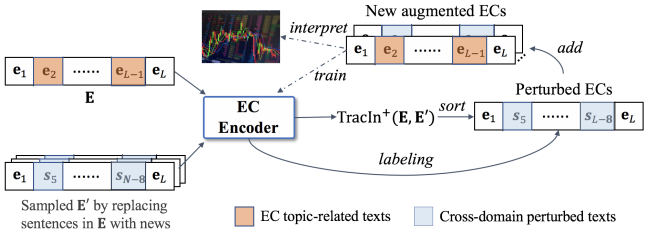}
    \caption{Unsupervised counterfactual augmentation.
    }
    \label{CL}
\end{figure}

\subsection{Multi-Domain Counterfactual Learning}

To overcome the above limitations, we further propose the multi-domain counterfactual learning framework, including the topic classifier and the unsupervised counterfactual augmentation block.

\subsubsection{Topic Classifier}
The topic classifier is devised for assigning each text to its matched topic, which can be treated as a lexical hub to build semantic connections between its covered texts from different domains.
A straightforward version of the topic classifier is Apache Solr search engine\footnote{https://solr.apache.org} that can implement full-text search with the queries of $N_f$ expert-defined topics.
However, its default algorithm BM25~\cite{robertson2009probabilistic} is a bag-of-words retrieval function, which can only rank the EC based on the exact appearance of topic terms rather than semantic similarity.
To improve semantic matching, we initialize the topic classifier as a pre-trained language model $LM(\cdot)$ with $N_f+1$ output label size, with an extra label indicating the non-topic alignment case.
Since the majority of top-$N_t$ sentences are semantically relevant to their assigned topics, we leverage the concept of distant supervision and treat them as positive samples and randomly sample $N_t$ irrelevant topic sentences as negative ones to fine-tune the topic classifier.
By assigning the topic to every EC and news sentence, we can pair up EC and news sentences by topic, to guarantee topic-wise semantic consistency.

\subsubsection{Unsupervised Counterfactual Augmentation}

As illustrated in Figure~\ref{CL}, following the concept of counterfactual learning (CL)~\cite{bottou2013counterfactual}, we further develop the unsupervised counterfactual augmentation block for (i) generating more perturbed EC instances for training, and (ii) investigating the attribution of critical EC contents w.r.t. the predicted volatility.
To ensure the effectiveness of cross-domain EC perturbation, we need to capture all potentially informative sentences by jointly considering their domain-topic relevance and task-aware significance.
Since the topic classifier bridges the topic-led connections among multi-domain texts, we next compare the task-aware salience between original EC sentences and perturbed ones to create the perturbations of similar task influence.
Specifically, we leverage the instance-based influence function~\cite{koh2017understanding} to encode the above comparison into the cross-entropy loss difference:
\begin{eqnarray}
\mathrm{pc}(\mathbf{E}^\prime|\mathbf{E}; \theta) = \mathcal{L}(\mathbf{E}^{\prime}; \theta) - \mathcal{L}(\mathbf{E}; \theta),
\label{celd}
\end{eqnarray}
where $\mathbf{E}$, $\mathbf{E}^{\prime}$ denote the embeddings of original and perturbed EC transcripts, respectively. 
Eq.~(\ref{celd}) is equivalent to $\mathrm{pc}(\mathbf{E}^{\prime}|\mathbf{E}, \theta) = \mathrm{logit}_{y}(\mathbf{E}; \theta) - \mathrm{logit}_{y}(\mathbf{E}^{\prime}; \theta)$, in which every term in the right hand side is the logit output evaluated at a model prediction $y$ from model $\theta$ right before applying the softmax function. 
Inspired by the classical result in the \cite{cook1982residuals}, we then quantify the model impact of $\mathbf{E}^{\prime}$ by adding a fraction of $\mathrm{pc}(\mathbf{E}^{\prime}|\mathbf{E}; \theta)$ scaled by a small value $\epsilon$ to the overall loss and obtain $\theta := \mathrm{argmin}_{\theta} \mathbb{E}_{\mathbf{E}_{i} \in \mathcal{E}^{train}} [\mathcal{L}(\mathbf{E}_{i}, \theta)] + \epsilon \mathcal{L}(\mathbf{E}^{\prime}; \theta) - \mathcal{L}(\mathbf{E}; \theta)$, where $\mathcal{E}^{train} \in \mathcal{E}$ denotes the EC set used for training.
The corresponding influence of up-weighing the importance of $\mathbf{E}^{\prime}$ on $\theta$ is:
\begin{eqnarray}
\frac{d \theta_{\epsilon, \mathbf{E}^{\prime}|\mathbf{E}}}{d\epsilon} \bigg\vert_{\epsilon = 0} = H^{-1}_{\theta}(\nabla_{\theta}\mathcal{L}(\mathbf{E}; \theta) - \nabla_{\theta} \mathcal{L}(\mathbf{E}^{\prime}; \theta)),
\end{eqnarray}
where $H_{\theta} = \frac{1}{|\mathcal{E}^{train}|}\sum_{\mathbf{E}_{i} \in \mathcal{E}^{train}}\nabla_{\theta}\mathcal{L}(\mathbf{E}_{i}, \theta)$ is the positive definite Hessian matrix by assumption.
By applying the above equation and the chain rule, we obtain a new influence closeness of $\mathbf{E}$ and $\mathbf{E}^{\prime}$ on $\theta$ by observing:
\small
\begin{align}
\nabla_\epsilon \mathcal{L} & (\mathbf{E}^\prime; \theta_{\epsilon; \mathbf{E}^{\top}|\mathbf{E}}) = \nonumber  \nabla_\theta \mathcal{L}(\mathbf{E}^\prime; \theta) H^{-1}_{\theta}(\nabla_{\theta}\mathcal{L}(\mathbf{E}; \theta) - \nabla_{\theta} \mathcal{L}(\mathbf{E}^{\prime}; \theta)).
\end{align}
\normalsize
If we assume that the influence of a training instance $\mathbf{e}$ is the sum of its contribution to the overall loss all through the entire training history, the above influence function can be relaxed into a form similar to the Hessian-free $\texttt{TracIn}^+(\cdot, \cdot)$~\cite{zhang2021sample} as: \small
\begin{eqnarray}
\texttt{TracIn}^+(\mathbf{E}, \mathbf{E}^\prime) = - \sum\limits_i 
\nabla_{\theta_i} \mathcal{L}(\mathbf{E}^\prime; \theta_i) \nabla_{\theta_i} \mathrm{pc}(\mathbf{E}^\prime|\mathbf{E}; \theta_i),
\label{tracin+}
\end{eqnarray}
\normalsize
where $i$ iterates through the checkpoints saved at different training steps.
The above equation intuitively measures the similarity of the model optimization direction between the entire perturbed EC and the perturbing sentence itself.
It also allows us to counterfactually evaluate how well the resulting EC transcript aligns with the perturbed sentence in terms of model impact, before using them for augmentation.

With the discovered informative contents, we further perform the unsupervised cross-domain perturbations for both the non-training-instance-based (NTIB) data augmentation and post-hoc explanation, which is more challenging yet crucial than conventional CL~\cite{dickerman2020counterfactual, verma2020counterfactual} that only relies on single-domain training instances.
More precisely, to augment a specific EC, each sentence is perturbed with $N_c$ randomly-chosen cross-domain sentences under a specific topic.
We calculate their corresponding $\mathrm{TracIn^{+}}$ scores, and view the top-$k_p$ and bottom-$k_n$ ranked sentences as the positively and negatively augmented instances, respectively.
Furthermore, the negatively-perturbed EC sentences are employed to provide domain-oriented explanations by quantifying the effect of a perturbed sentence on model prediction if the predicted volatility label is altered from right to wrong.
To smoothly optimize the encoder module, we directly assign the predicted labels to the positively-augmented ECs via the encoder, and the evenly-distributed target vector (i.e., (1/3, 1/3, 1/3)) to the negatively-augmented ones.

\noindent \textit{Remarks.} 
The above NTIB reasoning and augmentation are quite effective and robust because of two major reasons: 
(i) The task-adapted and domain-concerned properties of discovered EC explanations are ensured by the EC encoder and topic classifier, respectively;
(ii) The introduction of multi-domain semantics helps improve model robustness by amplifying the expressive manifold of informative contents.

\subsection{Model Training}
Overall, we define the following multi-class, cross-entropy loss for updating the EC encoder:
\begin{eqnarray}
\mathcal{L} = -\sum\limits_{c} p(c) \cdot \mathrm{log} \, q(c | \mathcal{F}(\mathbf{\tilde{E}})),
\end{eqnarray}
where $p(\cdot)$, $q(\cdot)$ are the label and the predicted score for each market volatility level $c \in \{0, 1, 2\}$, respectively.
We adopt a semi-supervised training protocol that alternates between the supervised learning of the EC encoder and the unsupervised augmentation using multi-domain data.
For the training data of the EC encoder, we use the original labeled EC pool in the first round, and separately add the newly-augmented data to the original pool in the subsequent rounds.
We repeat this iteration multiple times until converge.
Note that we enforce the predicted logits to be evenly distributed when the model is fed by the negatively perturbed data.
In addition, we introduce a simple yet effective bidirectional KL regularization trick~\cite{liang2021r} in the above loss, which enables the output distributions of different sub-models generated by dropout to be consistent with each other:
\small
\begin{eqnarray}
 \mathcal{L}_{KL} = \mathcal{L} +  \frac{\alpha}{2}\bigg[\mathcal{KL}(\mathcal{F}(\mathbf{\tilde{E}})_{1} || \mathcal{F}(\mathbf{\tilde{E}})_{2})) \nonumber  + \mathcal{KL}(\mathcal{F}(\mathbf{\tilde{E}})_{2} || \mathcal{F}(\mathbf{\tilde{E}})_{1})\bigg],
\end{eqnarray}
\normalsize
in which $\alpha$ is the hyperparameter, and $\mathcal{KL}(\cdot)$ is the Kullback-Leibler (KL) divergence between these two logits distributions upon dropout.
One remarkable finding is that two regularization techniques (i.e., KL divergence and data augmentation) used in \model~are empirically complementary to produce superimposed performances, which further highlights the significance of data augmentation to be applied in interpretable models. 

\section{Experiments}

\subsection{Experimental Setup}

\textbf{Data and Evaluation Metrics.}
We obtain $M = 17520$ quarterly ECs from WRDS\footnote{https://wrds-www.wharton.upenn.edu/} that includes 1,022 publicly traded US companies, ranging from May 1, 2018, to June 30, 2020. 
The multi-domain data includes the Earning Conference
Calls (short for QE) used in~\cite{qin2019you} and corporate financial news from Refinitiv\footnote{https://www.refinitiv.com/en/financial-news-services}.
The QE dataset covers 500 large public US firms during 2017, and the news covers all US company released news ranged from Oct 1, 2019, to Jan 1, 2020.
Each EC transcript contains the opening remarks (OP) session and the question-answering (QA) session, where we perform the session-based perturbation using either news or QE texts.
We select the $33^{rd}$ and $66^{th}$ percentiles of all log volatilities as thresholds for the three-class assignment.
We chronologically order and split the data into train, validation, and test based on the $8$:$1$:$1$ ratio.
The statistics of the datasets are summarized in Table \ref{datasets}.

We adopt a widely used metric \textit{Accuracy} for both short-term ($n = 3$) and long-term ($n = 15$) volatility prediction.

\begin{table}[t]
\centering
\Huge
\resizebox{0.48\textwidth}{!}{
\begin{tabular}{lrrr}
\hline
\textbf{Description} & \textit{Earnings call} & \textit{News sentence} & \textit{QE sentence} \\ \hline \hline
        $\#$ of involved companies & 1,022 & 4,266 & 576 \\
        $\#$ of total cases & 17,520 & 2,706,340 & 15,731,211\\
        $\#$ of total OP augmented cases & 17,520 & 2,622,897 & 83,443 \\
        $\#$ of total QA augmented cases & 17,520 & 7,789,062 & 7,942,149 \\
        \hline
        \end{tabular}
}
\caption{Statistics of datasets.}
\label{datasets}
\end{table}

\begin{table*}[t]
\setlength{\belowcaptionskip}{-10pt}
\centering
\Huge
\resizebox{0.85\textwidth}{!}{

\small
\begin{tabular}{c|c|ccc|ccc|ccc}
\hline
         \multicolumn{2}{c|}{\multirow{2}{*}{\textbf{Method}}} & \multicolumn{3}{c|}{Original EC} & \multicolumn{3}{c|}{News-augmented EC} & 
         \multicolumn{3}{c}{QE-augmented EC}\\\cline{3-11}
         \multicolumn{2}{c|}{} & OP & QA & OP + QA & aOP & aQA & aOP + aQA & aOP & aQA & aOP + aQA \\
        \hline \hline
         \multirow{10}*{n=3} & \multirow{1}*{RB} & 0.3380 & 0.3447 & 0.3142 & 0.3502 & 0.3417 & 0.3478 & 0.3661 & 0.3270 & 0.3289 \\
         & \multirow{1}*{TFB} & 0.3237 & 0.3237 & 0.3237 & 0.3237 &  0.3237 & 0.3237 & 0.3237 & 0.3237 & 0.3237  \\
         & \multirow{1}*{BOW} & 0.3679 & 0.3868 & 0.3545 & 0.3893 & 0.4021 & 0.3880 & 0.3160 & 0.3203 & 0.3111 \\ \cline{2-11}
         & \multirow{1}*{RoBERTa} & 0.2735 & 0.2698 & 0.2837 & 0.2764 & 0.2854 & 0.2909 & 0.2784 & 0.2646 &0.2937\\
         & \multirow{1}*{Longformer} & 0.3336 & 0.3191 &  0.3428 & 0.3567 & 0.3329 & 0.3510 & 0.3428 & 0.3573 & 0.3369 \\ 
         & \multirow{1}*{Informer} & 0.3583 & 0.3352 & 0.3649 & 0.3674 & 0.3418 & 0.3902 & 0.3567 & 0.3368 & 0.3711\\
        \cline{2-11}
         & \multirow{1}*{MDRM} & 0.4655 & 0.4570 &  0.3758 & 0.4839 & 0.4782 & 0.3959  & 0.4625 & 0.4411 & 0.3575 \\
         & \multirow{1}*{HTML} & 0.4204 & 0.3722 &  0.3734 & 0.4373 & 0.3734 & 0.3746 & 0.3673 & 0.3618 & 0.3716 \\
        \cline{2-11}
         & \model-P & 0.4997 & 0.4846 & 0.5012 & 0.5253 & 0.5032 & 0.5239 & 0.5272 & 0.4936 & 0.5147 \\   
         & \model & \textbf{0.5137} & \textbf{0.5102} & \textbf{0.5233} &  \textbf{0.5429} & \textbf{0.5328} & \textbf{0.5397} & \textbf{0.5368} & \textbf{0.5255} & \textbf{0.5271} \\ 
        \hline \hline
         \multirow{10}*{n=15} & \multirow{1}*{RB} & 0.3459 & 0.3584 & 0.3237 & 0.3454 & 0.3526 & 0.3347 & 0.3429 & 0.3386 & 0.3472 \\
         & \multirow{1}*{TFB} & 0.3237 & 0.3237 & 0.3237 & 0.3237 &  0.3237 & 0.3237 & 0.3237 & 0.3237 & 0.3237  \\
         & \multirow{1}*{BOW} & 0.3364 & 0.3493 & 0.3263 & 0.3601 & 0.3706 & 0.3884 & 0.2964 & 0.3014 & 0.2915 \\
        \cline{2-11}
         & \multirow{1}*{RoBERTa} & 0.2856 & 0.2734 & 0.2868 & 0.3012 & 0.2970 & 0.3051 & 0.2805 & 0.2659 & 0.2914 \\
         & \multirow{1}*{Longformer} & 0.2916 & 0.2818 &  0.3057 & 0.3287 & 0.3001 & 0.3162 & 0.3052 & 0.2966 & 0.3028 \\ 
         & \multirow{1}*{Informer} & 0.3383 & 0.3294 & 0.3445 & 0.3370 & 0.3328 & 0.3781 & 0.3354 & 0.3175 & 0.3452 \\
        \cline{2-11}
         & \multirow{1}*{MDRM} & 0.4329 & 0.4159 &  0.4103 & 0.4568 & 0.4461 & 0.4197  & 0.4560 & 0.4267 & 0.4072 \\
         & \multirow{1}*{HTML} & 0.4047 & 0.3916 &  0.3828 & 0.4255 & 0.4126 & 0.4052 & 0.3986 & 0.4092 & 0.4118 \\
        \cline{2-11}
         & \model-P & 0.4658 & 0.4404 & 0.4710 & 0.4964 & 0.4693 & 0.4746 & 0.4871 & 0.4692 & 0.4616 \\   
         & \model & \textbf{0.4719} & \textbf{0.4523} & \textbf{0.4830} &  \textbf{0.5059} & \textbf{0.4821} & \textbf{0.5053} & \textbf{0.4912} & \textbf{0.4793} & \textbf{0.5009} \\
        \hline
        \hline
        \end{tabular}
}
\normalsize
\caption{The accuracy performances of \model~together with all baselines on three datasets. The original EC dataset covers three cases of opening remark (OP), question answering (QA), and entire ECs (OP + QA), while each of two augmented datasets contains three cases of augmented OP (aOP), augmented QA (aQA), and augmented EC (aOP + aQA).
}
\label{overall}
\end{table*}

\noindent\textbf{Implementation Details.} 
Our model and all nine baselines are written in Pytorch, and trained by a single Tesla V100 GPU with the AdamW optimizer~\cite{loshchilov2017decoupled}.
We use the \textsf{RoBERTaForSequenceClassification}~\cite{liu2019roberta} as the pre-trained language model $LM(\cdot)$ to initialize all input embeddings and topic classifier.
For each EC sentence, we set the quantities $k_p$, $k_n$ of positively and negatively perturbed data to $1$.
The number of top queries to be considered in $SP(\cdot)$ is set to $10$, i.e., the average number of important sentences manually found in each EC transcript.
We create three different datasets for evaluation, including original ECs, news-augmented ECs, and QE-augmented ECs. 
Please refer to Appendix A for the details of network parameters.

\noindent\textbf{Baselines.} We compare our full approach with three fuzzy methods, three transformer-based methods, two EC-specific methods, and one variant of \model:

\begin{itemize}
    \item \textbf{Random baseline (RB)} is a uniform distribution inversion method that generates random numbers based on three possible outcomes.
    
    \item \textbf{Ticker following baseline (TFB)} treats the volatility of previous period as current prediction by assuming stationary market conditions.

    \item \textbf{Bag-of-words baseline (BOW)} \cite{sethy2008bag} encodes the EC transcripts into a bag of tokenized words or characters, which are fed into multi-layer perceptions for volatility modeling.

    \item \textbf{RoBERTa} \cite{liu2019roberta} develops an improved pre-training procedure to robustly optimize the standard BERT~\cite{devlin2018bert}. 

    \item \textbf{LongFormer} \cite{beltagy2020longformer} introduces a transformer-based network to combine the local-windowed and global attentions for efficient long-sequence modeling. 
    \item \textbf{Informer} \cite{zhou2021informer} proposes the ProbSparse self-attention to distill feature redundancy for handling the challenges of quadratic time complexity and memory usage in a vanilla transformer. 
    \item \textbf{MDRM}~\cite{qin2019you} devises a multi-modal deep regression model for volatility prediction.
    \item \textbf{HTML} \cite{yang2020html} develops a hierarchical EC-based transformer for multi-task volatility forecasting. 
     
    \item \textbf{\model-P} is a variant of \model~without the KL regularization term. 
\end{itemize}
For fair comparison, we fine-tune the model parameters, and set the iteration round to $2$, the max epoch of each round to be $40$ for all baselines. 
Note that we use the \textsf{RoBERTa-large}~\cite{liu2019roberta} to initialize the input embeddings of EC sentences for all the baselines except for RB and TFB.

\subsection{Overall Performance}
Table \ref{overall} reports the overall performance comparisons of our model \model~and all baselines. 
As can be seen, \model, together with its variants, significantly outperforms all other baselines on nine sub-datasets for (3-day, 15-day) volatility prediction.
We discover several interesting insights from the accuracy results:
\textbf{1)} EC encoder can empirically highlight the task-oriented importance of informative texts, making our \model~significantly outperform its closest transformer-based variants (e.g., HTML and Informer).
For example, by adding hierarchical sparse self-attention, \model~achieves ($9.33\%$, $6.72\%$), ($13.80\%$, $6.07\%$), ($14.99\%$, $10.02\%$), ($10.56\%$, $8.04\%$), ($15.94\%$, $6.95\%$), ($16.51\%$, $10.01\%$), ($16.95\%$, $9.26\%$), ($16.37\%$, $7.01\%$), ($15.55\%, 8.91\%$) improvements beyond another EC-based transformer with vanilla self-attention design, HTML, in average on all sub-datasets.
Besides, by removing the decoder part of a typical transformer, \model~improves Informer by ($15.54\%$, $13.36\%$), ($17.50\%$, $12.29\%$), ($15.84\%$, $13.85\%$), ($17.55\%$, $16.89$), ($19.10\%$, $14.93\%$), ($14.95\%$, $12.72$), ($18.01\%$, $15.58\%$), ($18.87\%$, $16.18\%$), ($15.60\%$, $15.57\%$) on nine sub-datasets.
The above results indicate that the EC encoder uses a lightweight structure to effectively distill redundant features and highlight the task-salient semantics for market volatility prediction.
\textbf{2)} The introduction of the unsupervised counterfactual augmentation supports most approaches to exhibiting better performances in multi-domain augmented datasets, where the model-based enhancements are ($0.79\%$, $0\%$, $2.96\%$, $1.39\%$, $1.85\%$, $1.28\%$, $2.05\%$, $1.39\%$, $2.17\%$, $2.57\%$), and ($0.43\%$, $0\%$, $-4.74\%$, $0.03\%$, $1.12\%$, $-0.13\%$, $-0.11\%$, $-0.41\%$, $1.51\%$, $1.77\%$) in average on two augmented datasets across both time spans, respectively.
Compared to QE-augmented data, the news-augmented one can provide more domain-relevant information gains for model update, thus achieving greater performance improvements.
\textbf{3)} EC, as a complicated long-sequence document, requires special design to achieve better understandings of its critical information.
This can be revealed from the observation that both EC-based baselines demonstrate obvious superiority over three generalized language models.
Finally, our full approach \model~outperforms all baselines by ($12.58\%$, $12.25\%$, $14.71\%$), ($14.19\%$, $13.59\%$, $14.85\%$), ($14.92\%$, $14.82\%$, $15.89\%$) in average on all three datasets across two different time periods.

\begin{figure}[t]
    \includegraphics[width = 0.48\textwidth]{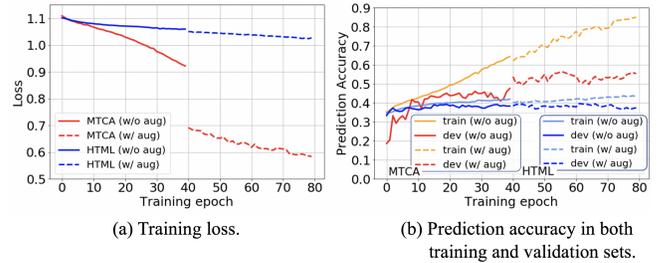}
    \caption{The convergence of two EC-based methods.}
    \label{rob}
\end{figure}

\begin{table*}[h]
\centering
\Huge
\resizebox{0.92\textwidth}{!}{
\begin{tabular}{c|c|c|c|c|c}
\hline
         \multirow{2}*{\textbf{Dataset}} &  \textbf{Negatively-perturbed EC} & 
         \textbf{Positively-related news} & \multirow{2}*{\textbf{Closeness}} & \textbf{Groundtruth} & \multirow{2}*{\textbf{Topic}} \\
         & \textbf{sentence as explanation} & 
         \textbf{for EC perturbation} & & \textbf{label} &\\
        \hline \hline
         \multirow{7}*{OP+News}
       
         & We were very much affected by  & market shares closed down 0.7\% due  & \multirow{2}*{0.5905} & \multirow{2}*{0} & \multirow{2}*{Workforce labour} \\
         & the truckers' strike & to the recent strike & & & \\ [3pt] \cline{2-6}
         
         & We are negotiating with our labor  & President of the powerful AFL-CIO labor  & \multirow{5}*{0.3085} & \multirow{5}*{0} & \multirow{5}*{Workforce labour} \\
         & union out here in the West, we're  & union gave the deal, his blessing and & & & \\
         & continuing to work with union leaders and & it's going to take a few years to begin  & & & \\
         & we look forward to reaching an agreement  &  to reverse the bad and the evil & & & \\
         & on that front. & that was done by NAFTA. & & & 
         \\ [3pt] \hline
         
         \multirow{7}*{QA+News} 
    
         & The other thing I think is just important  & MALLINCKRODT announces plans & \multirow{4}*{0.4026} & \multirow{4}*{2} & \multirow{4}*{Management pay} \\ 
         & to mention, on the incentive compensation for this & to update its incentive compensation & & & \\
         & quarter we have about 5 million roughly of additional & clawback policy and create & & &\\
         & incentive compensation related to the warrant gain. & OPIOID report & & &  \\[3pt] \cline{2-6} 
         
         & Unemployment is low, wages are   &  The cut would see his allowance  
         & \multirow{3}*{0.2966} & \multirow{3}*{0} & \multirow{3}*{Workforce pay} \\
         
         & rising and there's seemingly  & as a percentage of base salary drop to 15\%, & & & \\ 
         & no problem at all. & in line with the maximum contribution. & & & \\[3pt]
        \hline
        \end{tabular}
}
\caption{Qualitative study.}
\label{qanaly}
\end{table*}

\subsection{Ablation Studies of EC-based Approaches}
We further evaluate the effectiveness of our \model~and the closest baseline HTML in terms of their model convergences and performances. 
Specifically, for each model, we visualize its training loss, training accuracy, and validation accuracy in Figure~\ref{rob}. 
We adopt the one-round training iteration between the predictive module (e.g., \model's EC encoder, HTML) and the counterfactual-based generator module, where the predictive module takes 40 training epochs for both pre-and post-augmented modes.

\noindent\textbf{Convergences.}
Figure~\ref{rob}(a) shows that \model~converges much faster than HTML in both modes, probably because the hierarchical sparse self-attention empowers the EC encoder to precisely discover the topic-concerned contents that impose significant market effects.
We also observe a sudden loss drop that appears after \model~is switched to the post-augmented mode.
This could be explained by the effect of counterfactual augmentation, where the generator does amplify the expressive manifold of key EC contents and improve the model performance to a higher level.

\noindent\textbf{Performances.}
The detailed performance comparisons of these two EC-based methods are reported in Figure~\ref{rob}(b). 
Compared to HTML, our \model~achieves the best performances on two modes, in terms of both training and validation accuracies, which indicate the effectiveness of the proposed framework. 
Such learning superiority is also validated in Figure~\ref{rob}(a), where \model~achieves a better loss convergence rate.

\subsection{Qualitative Study}

Table~\ref{qanaly} showcases several news-supported examples of the negatively-perturbed instances for model explanations and the positively-perturbed ones for data augmentation in the 3-day volatility prediction task. 
Specifically, to understand model decisions, we first replace EC sentences with those negatively-related news contents of top-ranked $\mathrm{TracIn}^{+}$ scores, checking whether they have changed the volatility label predicted by the EC encoder.
Due to the space limit, we put one perturbed sentence of selected EC transcript in Table~\ref{qanaly},
where these perturbed examples are found to provide strong evidence for the predicted label.
For instance, the EC statement "We were very much affected by truckers’ strike" conveys the management's negative expectations on corporate situations, which partially accounts for the downward market trend, labeled as 0. 
As can be seen, our unsupervised counterfactual augmentation successfully quantifies the degree of domain-topic relevance and task-based salience for EC informativeness modeling, which helps discover the decision-influencing EC texts as post-hoc interpretations.
In practice, these sentences do not completely lead to actionable insights because market movement also heavily depends on other important EC-undeclared factors (e.g., management turnover~, competitor analysis), yet offer domain experts with an accurate picture of informative EC contents for reference and in-depth analysis.

Second, we demonstrate some cases of positively-related news perturbations in Table~\ref{qanaly}, which not only validate the task-aware significance of perturbed data using $\mathrm{TracIn}^{+}$ function, but also virtually guarantee the multi-domain semantic consistency using topic classifier during data augmentation.
An interesting finding is that compared to OP-based augmentation, the QA-based one tends to be semantically distant from original EC texts based on the token-level closeness, which thereby explains the relatively low-performance enhancements on most sub-datasets, as reported in Table~\ref{framework}.
For example, the EC statement "Unemployment is low, wages are rising and there’s seemingly no problem at all" delivers the opposite semantics compared to its positively-related news.
It is probably because numerous QA contents are long informal narratives and can hardly match those refined news statements based on the token-level retrieval function (e.g., BM25 algorithm).

\section{Conclusion and Future Work}

In this paper, we proposed a multi-domain transformer-based counterfactual augmentation (\model) framework for interpretable EC-based financial application.
Specifically, we first developed the EC encoder,
where the hierarchical sparse self-attention models the sentence-level long-range dependencies and highlights the task-inspired importance of informative texts for market inference.
Then, we introduced the multi-domain counterfactual learning, where the topic classifier achieves the topic-text alignment to build the topic-led connections among multi-source texts.
We later employed the unsupervised counterfactual augmentation block to quantify the gradient-based variations of perturbed ECs, which allows \model~to perform CL-based augmentation and provide the NTIB explanations.
Finally, extensive experiments demonstrated that our \model~can precisely forecast and reason upon future market uncertainty.

This work is one step towards long-document modeling for financial tasks based on EC transcripts.
Future research in this area may include the following: (i) investigating other lengthy documents for volatility prediction, such as industrial report, 10-K form.
(ii) considering other real-world financial applications, including stock trading, portfolio management, and default risk evaluation.

\bibliography{reference}
\appendix

\newpage
\section{Appendix A: Implementation Details}
We fine-tune the parameters of topic classifier based on its last two layers and output head using a batch size of $64$ and max epochs of $100$.
We set the learning rate to $0.00001$, the batch size to $64$, the dimension $d$ of all latent representation vectors to $512$, the dropout rate to $0.2$, the maximal sentence length of EC  $L$ to $500$, the number $H$ of sparse self-attention heads in $[\cdot]_{HS}$ function to $8$, the activation function in $SP(\cdot)$ to Parametric ReLU, the number $N_f$ of expert-refined topics to $221$, the weight decay $\gamma$ to $0.01$, the number $N_c$ of randomly-chosen cross-domain sentences to $10000$, the number $N_t$ of top-ranked sentences used for topic classifier to $10$, and the hyperparameter $\alpha$ in $\mathcal{L}_{KL}$ to $0.3$. 
Note that the implementation code of our \model~is included in the supplementary materials.

\end{document}